\documentclass[conference]{IEEEtran}
\IEEEoverridecommandlockouts
\usepackage{cite}
\usepackage{amsmath,amssymb,amsfonts}
\usepackage{algorithm}
\usepackage{algorithmic}
\usepackage{graphicx}
\usepackage{textcomp}
\usepackage{xcolor}
\def\BibTeX{{\rm B\kern-.05em{\sc i\kern-.025em b}\kern-.08em
    T\kern-.1667em\lower.7ex\hbox{E}\kern-.125emX}}
\begin{document}
\setcounter{secnumdepth}{4}
\setcounter{tocdepth}{4}
\title{A Support Vector Model of Pruning Trees Evaluation Based on OTSU Algorithm\\
{\footnotesize}
\thanks{This research is funded by Natural Science Foundation of Zhejiang Province (Grant No. LQ20G010002), and the National Science Foundation of China (71571162, 71702164). This research is supported by the project of China (Hangzhou) cross-border electricity business school, the Philosophy and Social Science Foundation of Zhejiang Province (21NDJC083YB). .}
}

\author{
\IEEEauthorblockN{Yuefei Chen}
\IEEEauthorblockA{
\textit{Rutgers University} \\
Piscataway, NJ, USA \\
yc1023@scarletmail.rutgers.edu}
\and
\IEEEauthorblockN{Xinli Zheng}
\IEEEauthorblockA{
\textit{Zhejiang Tianyanweizhen}\\ \textit{network Polytron Technologies}\\
Hanzhou, Zhejiang, China \\
email address or ORCID}
\and
\IEEEauthorblockN{Chunhua Ju}
\IEEEauthorblockA{\textit{Contemporary Business}\\ \textit{and Trade Research Center} \\
\textit{Zhejiang Gongshang University}\\
Hangzhou, Zhejiang, China \\
 Jch@zjgsu.edu.cn}
\and
\IEEEauthorblockN{Fuguang Bao}
\IEEEauthorblockA{\textit{School of Management Science and Engineering} \\
\textit{Zhejiang Gongshang University}\\
Hangzhou, Zhejiang, China \\
baofuguang@126.com}

}

\maketitle

\begin{abstract}
The tree pruning process is the key to promoting fruits' growth and improving their productions due to effects on the photosynthesis efficiency of fruits and nutrition transportation in branches. Currently, pruning is still highly dependent on human labor. The workers' experience will strongly affect the robustness of the performance of the tree pruning. Thus, it is a challenge for workers and farmers to evaluate the pruning performance. Intended for a better solution to the problem, this paper presents a novel pruning classification strategy model called “OTSU-SVM” to evaluate the pruning performance based on the shadows of branches and leaves. This model considers not only the available illuminated area of the tree but also the uniformity of the illuminated area of the tree. More importantly, our group implements OTSU algorithm into the model, which highly reinforces robustness of the evaluation of this model. In addition, the data from the pear trees in the Yuhang District, Hangzhou is also used in the experiment. In this experiment, we prove that the OTSU-SVM has good accuracy with 80\% and high performance in the evaluation of the pruning for the pear trees. It can provide more successful pruning if applied into the orchard. A successful pruning can broaden the illuminated area of individual fruit, and increase nutrition transportation from the target branch, dramatically elevating the weights and production of the fruits.
\end{abstract}

\begin{IEEEkeywords}
pruning, classification strategy, support vector machine model, OTSU Algorithm
\end{IEEEkeywords}
\section{Introduction}
\label{sec:intro}
With the rapid development of modern agriculture, an increasing number of farmers tend to grow crops technically and scientifically. As a significant role in the growth of the fruit tree, the pruning process is essential for improving the production of the fruit. Proper training through correct pruning is important for a healthy, strong fruit tree \cite{feucht2000training}. Generally, pruning will be done during the summer. It is the best way to remove vigorous, upright water sprouts so that overly-vigorous trees can be brought into balance again. 

Considering the relationship between pruning and production, three factors of pruning will affect the production of fruit. At first, pruning can leave enough space for fruit growth, in case that a lot of small fruits crowd in a small space and compete for resources like air, water transportation. A crowded space will restrain respiration and transpiration, which can speed up fruits decaying and influence the quality. More importantly, it will not stimulate the new branch growth which is not desired \cite{maughan2017training}. A new branch represents a new competitor for the nutrition and organism. The branch with fruit will obtain fewer resources than what is obtained in the situation without a new branch. Thus, pruning can ensure each branch enough transportation of nutrition and organism. Additionally, pruning can expand the overall light area and accelerate transmittance throughout the canopy of the tree. It can increase the amount of light by approximately 25$\%$ in the lower part of the canopy compared to the group without pruning controlled \cite{bhusal2017summer}. With the purpose to increase the production of fruit, light intensity and luminance are ignorable issues to be concentrated on. These factors will affect the absorbance of lights and the photosynthesis efficiency of fruits. Yuan Gao and his team has presented the light distribution of the tree is highly correlated with the growth of fruit. High light can promote photosynthetic ability thereby effectively decreasing fruit abscission in fruit trees \cite{gao2018canopy}.

Currently, there are two kinds of methods for pruning fruit trees -- manual pruning and automation technologies. The majority of fruit tree pruning systems are highly dependent on human labor. Experienced laborers are in great demand for orchard pruning, who can make a highly successful rate of pruning based on their experience. What they do can promote fruit production. However, with the increasing cost of labor and limited labor availability, it is highly desirable for growers to seek mechanical or robotic solutions for performing these production operations \cite{zahid2020development,he2018sensing}. Meanwhile, the accuracy of pruning becomes an unignorable factor for researchers when they are designing automated pruning systems because only successful pruning can promote the increment of production. In addition, mechanical pruning had a significantly negative impact on the photosynthetic rate \cite{lauvzike2020impact}. Thus, researchers have to evaluate the performance of fruit tree pruning and make sure of the robustness of this accuracy in the automated pruning systems.

Therefore, in this paper, we propose a novel pruning evaluation model based on the shadow of the branches and leaves. To our knowledge, there is little research aiming at evaluating pruning performance by using the shadows of the fruit trees. We explore this aspect and develop the model for pruning. The key contributions of this paper are as follows.
\begin{itemize}
\item The paper conducts a comprehensive investigation to study the current models which predicts the pruning process and related pruning strategies, and we also go to the orchard to take the images and discuss with the agricultural experts. Our investigation drives us to create the dataset including the information about the fruit trees and their pruning performance. Based on this dataset, we analyze those two features of the pruning shadows, which significantly affect the pruning performance and production of the fruit trees.
\item Additionally, the paper proposes a new simple model called OTSU-SVM to evaluate the pruning performance. In contrast to the model which uses deep learning like back propagation neural network \cite{LiXinxing2021Agricultural},  this model does not need a great scale of data. The OTSU-SVM will select and extract the features of the shadows of fruit trees and preprocess them. These results will be used to train a supported vector machine (SVM) model to classify the different performances of pruning.
\item More importantly, the paper extensively implements and evaluates OTSU-SVM based on the real-world data in Hangzhou including the shadows of pear trees in the orchard and pruning performance. The result shows that the accuracy of OTSU-SVM can go to 80$\%$ and give most of the shadows of the fruit trees a proper pruning evaluation. 
\end{itemize}

The rest of the paper is organized as follows. Section~\ref{sec:related} introduces the related work. In Section~\ref{sec:method}, we depict the details about the methods and materials about the OTSU-SVM model. Section~\ref{sec:evaluation} presents the process of our experiment in the orchard and Section~\ref{sec:experiment} illustrates the result of the experiment of our OTSU-SVM model and comparisons with another basic model. Section~\ref{sec:conclusion} displays a discussion about the result and interprets the findings, and then we conclude this paper and discuss the future work.

\section{Related Work}
\label{sec:related}
\subsection{Automation Pruning System}
Pruning is highly dependent on human labor. However, skilled labor is in short supply, and the increasing cost of labor is becoming a big issue for the tree industry. Meanwhile, work safety is another issue in the manual pruning \cite{he2018sensing}. To address these challenging pruning issues, lots of researchers have concentrated on the alternate aspect about automation pruning system. Some of them tend to develop an application with knowledge of complex skills like tree training and pruning to teach growers how to evaluate it \cite{kolmanivc2017computer}. Additionally, many scholars use detection and 3d reconstructions to analyze the tree construction and make decisions \cite{zahid2020development,he2018sensing,LiXinxing2021Agricultural}. Most of them apply computer vision and 3d modeling technology to reconstruct the branches of trees. However, they have not considered the effect of leaves of trees on the fruits, which results in the factor from the light and photosynthesis efficiency of fruits is ignored so that the strategy of pruning will be involved. What motivates us to explore a new mechanism considering the factors of leaves and the light area of fruits.
\subsection{Pruning Decision Model}
The pruning decision model becomes a popular research topic in recent years with the development of robotic and automatic pruning. Barbosa and his team focus on the pruning strategies on pear trees with phenological evaluation. They propose to analyze the variation of climate in a year and fruit habits in different cultivars and then find how to affect the determination of pruning strategies \cite{de2018phenological}. Dr. Schupp and his team proposed a severity level pruning strategy for tall spindle apple trees to provide guidelines for determining the cutting threshold for robotic pruning \cite{schupp2017method}. 
\subsection{Summary}
Technically, in contrast to the other pruning model, the key advantage of the method in this paper is that authors consider the effects of leaves and light area of fruits, which is also a significant factor of the production of the fruit, besides the effect of branches. More importantly, the model does not need a large scale of data for training driving and is not complex as the model that needs 3-d reconstruction \cite{LiXinxing2021Agricultural} but still has a high performance of evaluation on fruit tree pruning.
\section{Method}
\label{sec:method}
In this section, the pruning data is firstly introduced. After that, an overview of the model “OTSU-SVM” is given and illustrated into two parts of the model in detail.
\subsection{Data Descriptions}
\begin{figure}[htbp]
\centerline{\includegraphics[scale = 0.4]{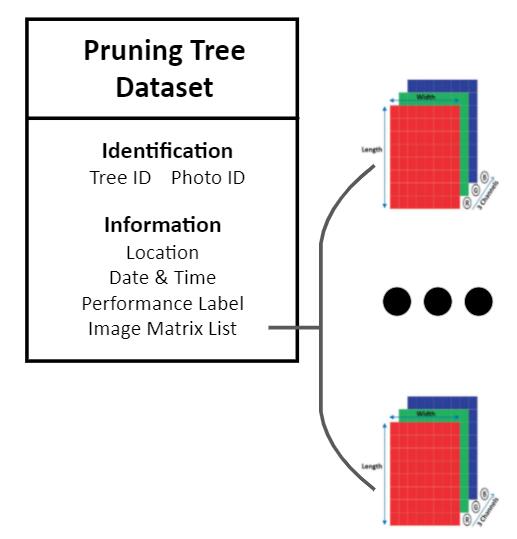}}
\caption{The metadata of pruning tree dataset}
\label{fig1}
\end{figure}
\begin{figure}[htbp]
\centerline{\includegraphics[scale = 0.65]{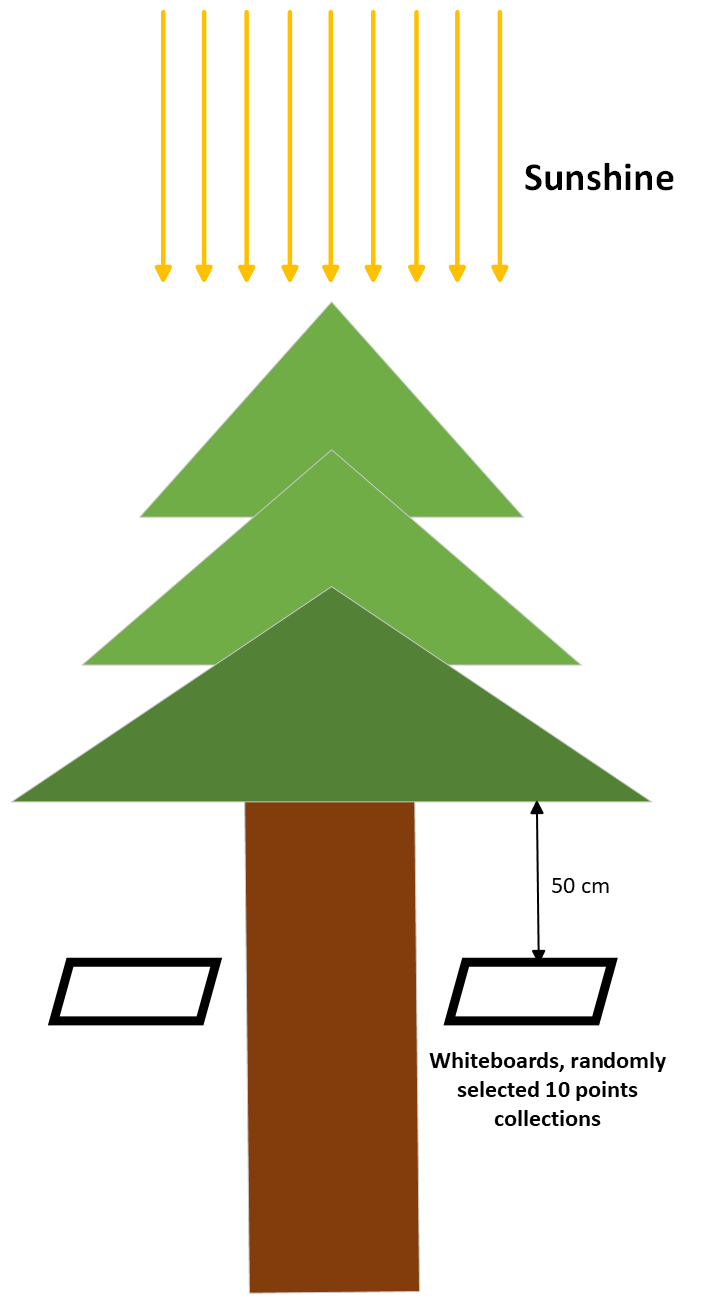}}
\caption{collection of shadows under the pear trees}
\label{fig8}
\end{figure}
\begin{figure}[htbp]
\centerline{\includegraphics[scale = 0.5]{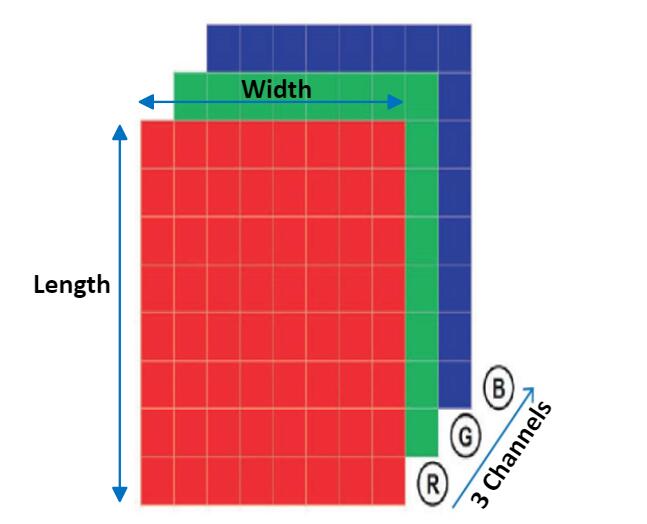}}
\caption{The metadata of pruning tree dataset}
\label{fig2}
\end{figure}

Pear trees are traditional and prevailing trees growing in China. In Hangzhou, there are thousands of hectares of pear trees. It is convenient for researchers to collect the data and store it in the pruning dataset. In the dataset, the metadata of the dataset is shown in Fig.~\ref{fig1}. Every tree has its id information and individual photo id. Then, they have all labeled the pruning performance which is evaluated by experts. $1$ represents performs good and $0$ represents perform not good. Shadows of the pear trees are collected in the form of images. The process of collection is shown in Fig.~\ref{fig8}. 
The collection time is at noon and the sun light is nearly vertical to the ground. The white board is used to get shadow of the leaves and branches. The distance between whiteboard and branches are $0.5$ meter. With the expansion of the branches of trees, ten points will be randomly selected under the tree to take images. When they are stored in the database, it will be transformed into a 3-dimensions of images matrix as shown in Fig.~\ref{fig2}. It is the $3 \times n \times m$ matrix where n and m are the length and width of the image. The other dimension means three channels of one pixel of RGB image, where $Image[x][y] = (r, g, b)$ indicates that the pixel at the position $(x, y)$ has a color $(r, g, b)$. In this pixel, $r$ represents the level of red, $g$ represents green, and $b$ represents blue. The number of the matrix as values between $0$ and $255$ represents the intensity of the corresponding channel.
\begin{figure}[htbp]
\centerline{\includegraphics[scale = 0.5]{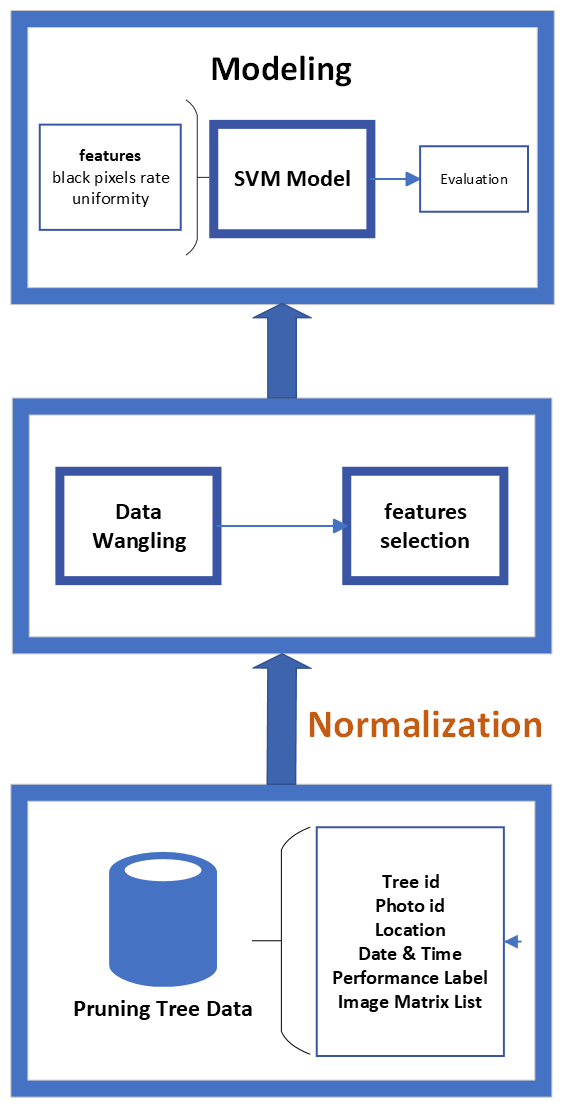}}
\caption{Three layers of feature recognition architecture}
\label{fig3}
\end{figure}
\subsection{Overviews of OTSU-SVM}
The OTSU-SVM model is designed to analyze the image data from the fruit tree. It comprises three parts. As shown in Fig.~\ref{fig3}, a three-layer feature recognition architecture called OTSU-SVM for fruit trees is presented. The bottom layer is the image data layer. It collects and stores data from the orchard and lays the data foundation for feature selection and model evaluation. The second layer is the data wrangling and feature selection layer. This layer preprocesses the image data and normalizes the image matrix. Additionally, the features are also selected and analyzed, which is the input for the next layer. In the top layer, i.e., Support Vector Machine (SVM) model. We use the features obtained from the data wrangling and feature selection layer to classify the points and train the model. After training the model, it can evaluate the pruning performance of the fruit trees and classify which points of fruit trees perform well.

\subsubsection{Data Layer}
The data in the dataset is the image data of each selected fruit tree. Its details are given in the data description section. These data including the shadows distribution of the fruit trees.
\subsubsection{Data Wrangling and feature selection Layer}
\begin{figure}[htbp]
\centerline{\includegraphics[scale = 0.3]{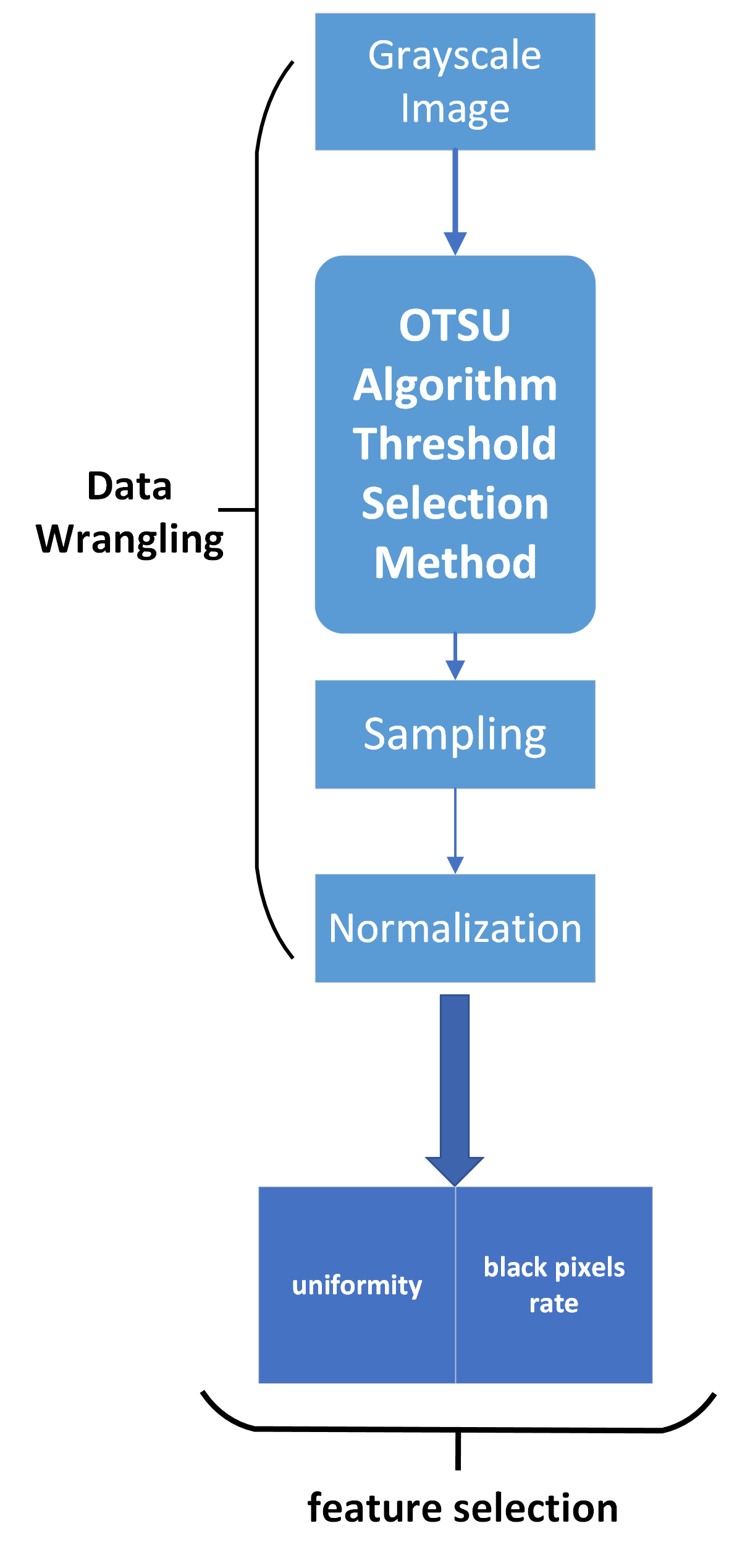}}
\caption{The layer of preprocessing}
\label{fig7}
\end{figure}

In this layer, there are two parts of the preprocessing. The details are shown in Fig.~\ref{fig7}. The first part is data wrangling and contains four steps. In these steps, the first step is to transform the 3 channels RGB image into 1 channel grayscale image and store it into an image matrix. The second step is the significant part in the data wrangling. It uses OTSU algorithm to set threshold of two classes, dark and white pixels, and separate gray scale pixels into these two classes. The OTSU algorithm is an effective threshold selection method and seeks the threshold which can lead to a high accuracy of analysis. Additionally, the next step is sampling. Sampling is optional in this research. That is because it aims to reduce the complexity of the matrix and lessen the burden of the computer. if the research group has high performance computing capability, this step is recommended to skip because skipping this step can help obtain more accurate result. At last, the data needs to be normalized into the range from $0$ to $1$. The purpose is to decrease the deviation and it is prepared for the feature selection.

The second part is feature selection. In this part, uniformity and black pixels rate are selected as the feature of image matrices.
\begin{figure}[htbp]
\centerline{\includegraphics[scale = 0.3]{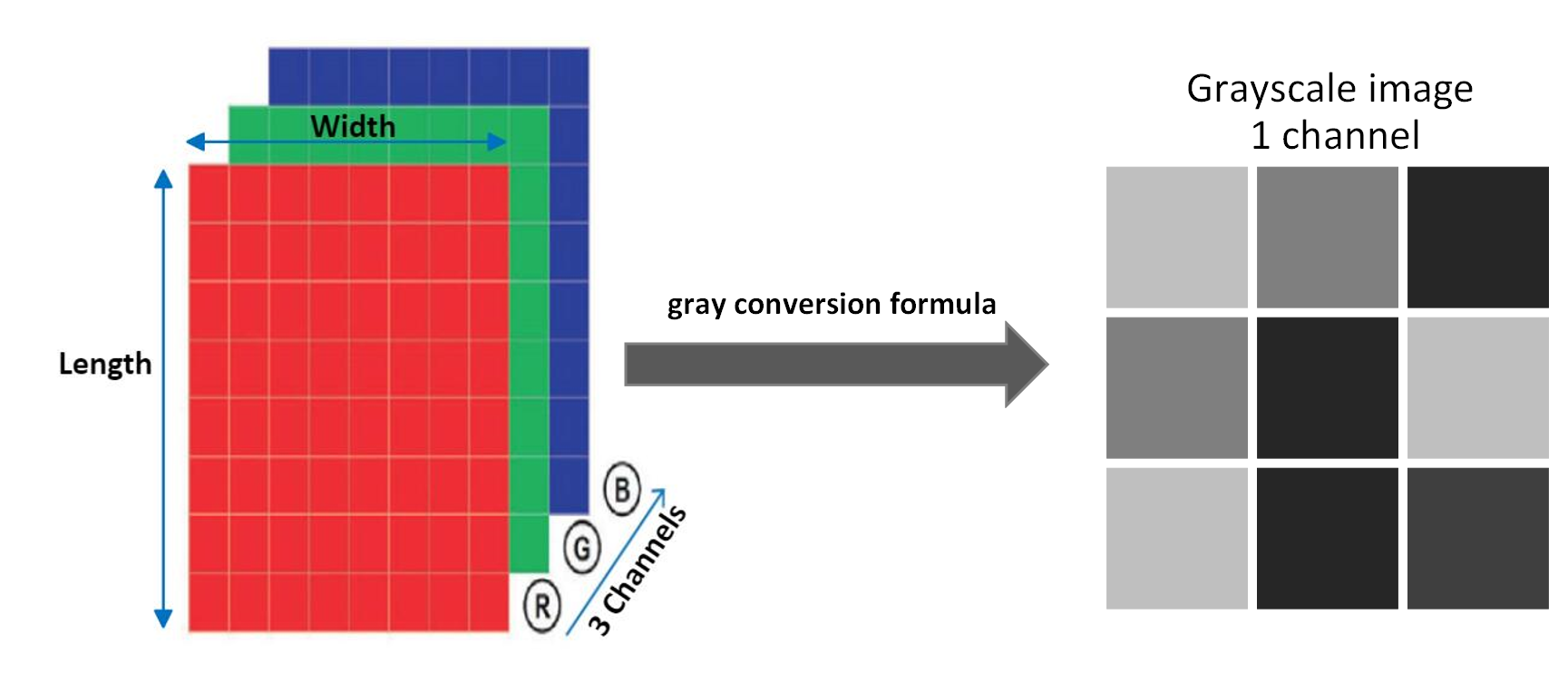}}
\caption{The 3 channels image conversion into 1 channel grayscale image}
\label{fig17}
\end{figure}
\paragraph{Data wrangling}

The paper explores data from the bottom layer to extract the features of the image of shadow. In order to fulfill the purpose, the 3 channels RGB image matrix is transformed into 1 channel grayscale image. A classic RGB image to gray conversion formula is given by 
\begin{equation}
Gray (r,g,b) = 0.21r + 0.72g + 0.07b\label{eq1}
\end{equation}
In this equation, the resulting value $Gray (r, g, b)$ is between 0 and 255 and $r, g, b$ represents the levels of red, green, and blue. The Fig.~\ref{fig17} shows that process of the conversion.

Moreover, the grayscale image also has too much-complicated data to process. The purpose of paper is to distinguish the part of the shadow and the area of lights. Thus, only two values of the data is in demand and the shadow pixel is set to be $0$, which represents the color of the pixel is black, and the light pix as $255$ and white. To separate these two parts of the image, thresholding is an optimizing way to solve it. The threshold selection method called OTSU algorithm \cite{otsu1979threshold} with our statements of $0$ and $255$ is used to set a threshold and divide the pixels of the image into two parts. The OTSU algorithm is used to perform automatic image thresholding in computer vision and image processing \cite{sezgin2004survey}. In this paper, the algorithm returns a intensity threshold that separate pixels into two classes, shadows and lights. This threshold is determined by minimizing intra-class intensity variance, or equivalently, by maximizing inter-class variance \cite{otsu1979threshold}. The intra-class variance is defined as a weight sum of variance of separated two classes,
\begin{equation}
\sigma^2_w(t) = w_0(t) \sigma^2_0(t) + w_1(t) \sigma^2_1(t)\label{eq2}
\end{equation}
In this equation, weights $w_0$ and $w_1$ are the probabilities of the two classes separated by a threshold $t$, and $\sigma^2_0$ and $\sigma^2_1$ are variance of these two classes. The class probability $w_0(t)$ and $w_1(t)$ can be expressed as, 
\begin{equation}
\begin{split}
w_0(t)=\sum^{t-1}_{i=0}p(i),\\[1mm]
w_1(t)= \sum^{L-1}_{i=t}p(i)\label{eq3}
\end{split}
\end{equation}
In this equation, the range is between $0$ and $L-1$. In addition, the inter-class variance is defined as the variance between two separated classes. The equation is 
\begin{equation}
\begin{split}
\sigma^2_b(t) = \sigma^2 - \sigma^2_w(t) & = w_0 (\mu_0 - \mu_T)^2 + w_1 (\mu_1 - \mu_T)^2  \vspace{1ex}\\
                                        & = w_0(t) w_1(t) [\mu_0(t) - \mu_1(t)]^2\label{eq4}
\end{split}
\end{equation}
In this equation, $\mu$ is the class mean of these classes. The class means $\mu_0(t)$, $\mu_1(t)$ and $\mu_T$ are expresses as, 
\begin{equation}
\begin{split}
\mu_0 = \frac{\sum^{t-1}_{i=0}ip(i)} {w_0(t)} \vspace{1ex},\\[1mm] \mu_1 = \frac{\sum^{L-1}_{i = t} ip(i)} {w_1(t)} \vspace{1ex},\\[1mm] \mu_T = \sum^{L - 1}_{i = t}ip(i)\label{eq5}
\end{split}
\end{equation} 
According the equation \eqref{eq4}, the relationship between the inter-class variance and the inter-class variance can be verified that minimizing intra-class variance is equivalent to maximizing inter-class variance.The algorithm 1 with our statement is as follows.

\begin{algorithm}
    \renewcommand{\algorithmicrequire}{\textbf{Input:}}
    \renewcommand{\algorithmicensure}{\textbf{Output:}}
    \caption{OTSU thresholding algorithm with statements}
    \begin{algorithmic}
        \REQUIRE Image Matrix $S$ with length $m$ and width $n$
        \ENSURE Thresholding Matrix $S_1$
        \STATE Define $min \ variance = \infty$
	    \STATE Define $threshold = -1$
	    \STATE Define $p = 0$
	    \WHILE{$p<256$}
	    \STATE Define $List_A, List_B$ as empty lists
	    \STATE Define $i, j = 0$
	    \WHILE{$i<m$}
	    \WHILE{$j<n$}
	    \IF{$S[i][j] \geq p:$}
	    \STATE $List_A.append(S[i][j])$
	    \ELSE 
	    \STATE $List_B.append(S[i][j])$
	    \ENDIF
	    \ENDWHILE
	    \ENDWHILE
	    \STATE $S_A = var(A)$
		\STATE $S_B = var(B)$
		\STATE $I_A = sum(A)$
		\STATE $I_B = sum(B)$
		\STATE $Intra\ class\ variance = S_A \times I_A + S_B \times I_B$
        \IF{$Intra\ class\ variance < min\ variance$}  
		\STATE$min variance = Intra class variance$
		\STATE$threshold = p$
        \ENDIF
	    \ENDWHILE
	    \STATE Define $S_1$ as empty array with length $m$ and width $n$
	    \STATE Define $k, l = 0$
	    \WHILE{$k<m$}
	    \WHILE{$l<n$}
	    \IF{$S[k][l] \geq threshold$}
	    \STATE $S_1[i][j] = 255$
	    \ELSE 
	    \STATE $S_1[i][j] = 0$
	    \ENDIF
	    \ENDWHILE
	    \ENDWHILE
        \RETURN $S_1$
    \end{algorithmic}

\end{algorithm}

\begin{figure}[htbp]
\centerline{\includegraphics[scale = 0.8]{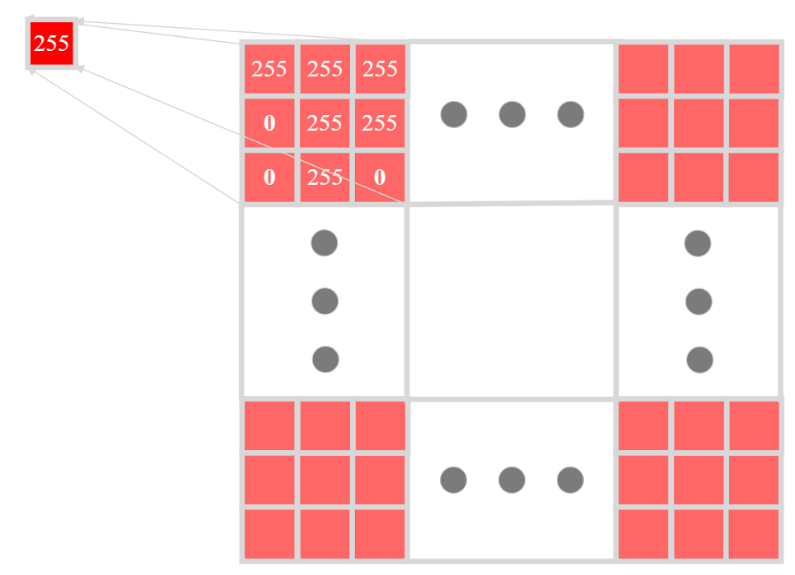}}
\caption{The selection process of the represented values of the image}
\label{fig4}
\end{figure}

Additionally, as an image matrix, its width and length are generally over than 2000 pixels. In other words, mostly image matrices contain more than four million pixels. It will leave a heavy burden to the program and the CPU computing. The algorithm is designed to sample the image matrix by summarizing the colors of pixels in patches of the matrix. The algorithm is as follows. 
\begin{algorithm}
    \renewcommand{\algorithmicrequire}{\textbf{Input:}}
    \renewcommand{\algorithmicensure}{\textbf{Output:}}
    \caption{Sample image matrices algorithm}
    \begin{algorithmic}
        \REQUIRE Image Matrix $S$ with length $m$ and width $n$, shrinking times $p$
        \ENSURE Adjusted Matrix $S_1$
        \STATE Define $S_1$ as empty array with length $m$ and width $n$
        \STATE Define $i, j = 0$
	    \WHILE{$i<m/p$}
	    \WHILE{$j<n/p$}
	    \STATE $Sum = sum(S[p * i][p * j] + … + S[p * (i - 1) + 1][p * j] + … + S[p * (i - 1) + 1][p * (j - 1) + 1] )$
	    \IF{$Sum < 5$}
	    \STATE $S_1[i][j] = 0$
	    \ELSE 
	    \STATE $S_1[i][j] = 255$
	    \ENDIF
	    \ENDWHILE
	    \ENDWHILE
	    \RETURN $S_1$
    \end{algorithmic}

\end{algorithm}
In details, this process is similar with the pooling layer \cite{yu2014mixed} in the deep learning and convolutional neural network (CNNs) \cite{albawi2017understanding,kim2017convolutional}. The algorithm aims to find the representative value in a $p \times p$ grids as its value and summarize the image matrix. It is to progressively reduce the spatial size of the representation to reduce the amount of computation in the following normalization and model. In this experiment, the p is set to be 3. The Fig.~\ref{fig4} shows that process of selecting the represented values of the image. In other words, the size of the new image matrices is diminished into $1/9$ of the original image matrices, which can improve computation efficiency and does not lose a lot of information about the image. 
\paragraph{features selection}

\begin{figure}[htbp]
\centerline{\includegraphics[scale = 1.25]{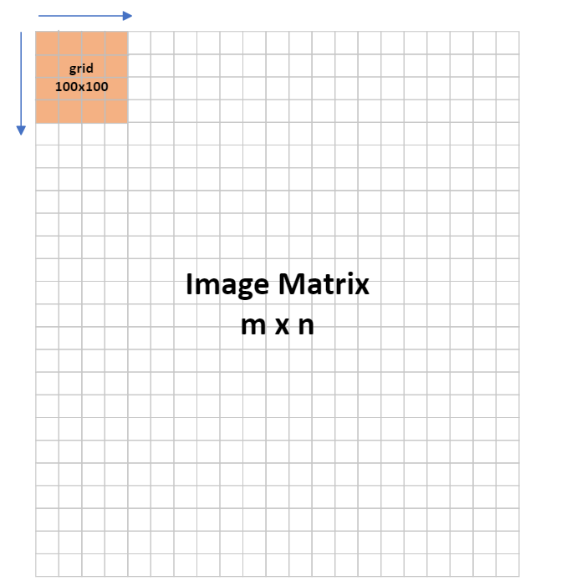}}
\caption{A grid moving and counting the number of white pixels in the grid}
\label{fig5}
\end{figure}

More importantly, we are based on the preprocessed data to extract features. We consider the lights area of the fruit and the effects of the distribution of the leaves and branches. Therefore, the shadow of the fruit tree is a good choice to analyze these factors. We find two features from the shadow. The one is black pixels rate. The other one is uniformity. The black pixels rate means the rate of black pixels in the whole image. The equation is 
\begin{equation}
\begin{split}
black\ pixels\ rate = \frac{count(black pixels)}{count(all pixels)}
\end{split}
\end{equation}
It reveals the light transmittance and the light area of the fruit tree. If the black pixels rate is close to $1$, the transmittance of sun light is close to $0$. The uniformity is an indicator that can determine if the distribution of the branch and leaves is uniform. The approach to getting the uniformity is as follows. In this experiment, the size of a grid is set  to be $100 \times 100$. In detail, the number of white pixels are counted in every grid of the matrix and store into the list. The Fig.~\ref{fig5} shows the process of the grid moving and counting the number of white pixels. 

After that, the variance of the values is calculated in the list to get the standard deviation of these grids. Standard deviation in these grids represents the extent of deviation between these grids. When these grids deviation are large, it means they are not uniform. If it is close to 0, the matrix is uniform. Thus, the standard deviation is equivalent to the uniformity we defined. The equation is as follows.
\begin{equation}
\begin{split}
uniformity = \sigma = \sqrt{\frac{\sum{(x_i-\mu)^2}}{N}}
\end{split}
\end{equation}
In this equation, $N$ is the number of grids. $x_i$ is the number of white pixels in each grid. $\mu$ is the mean of $x_i$ in all grids. Using this equation, we build the matrix of uniformity in this algorithm 3.  
\begin{algorithm}
    \renewcommand{\algorithmicrequire}{\textbf{Input:}}
    \renewcommand{\algorithmicensure}{\textbf{Output:}}
    \caption{calculate the uniformity}
    \begin{algorithmic}
        \REQUIRE Processed Image Matrix $S$ with length $m$ and width $n$, grid $G$ with size of $p \times p$ 
        \ENSURE uniformity

        \STATE Define empty list $L = []$
        \STATE Define $i, j = 0$
	    \WHILE{$i<m/p$}
	    \WHILE{$j<n/p$}
	    \STATE $Sum = sum(S[p * i][p * j] + … + S[p * (i - 1) + 1][p * j] + … + S[p * (i - 1) + 1][p * (j - 1) + 1] )$
	    \STATE $L.append(Sum)$
	    \ENDWHILE
	    \ENDWHILE
	    \STATE $uniformity = \sqrt{(variance(L))}$
	    \RETURN $uniformity$
    \end{algorithmic}
\end{algorithm}

In addition, the ranges of the two features are different. In other words, different ranges will have unstable effects on the weights of two variables in the SVM model. Therefore, it is necessary for the model to normalize two features into the data with a range from 0 to 1. The normalization is frequently applied in image recognition \cite{pei1995image,alghoniemy2000geometric,abu1985image,koo2017image,loizou2009brain}. The goal of the normalization is to transform features to be on a similar scale and improves the performance and training stability of the model. There are four common normalization techniques. That is scaling to a range, clipping, log scaling, and z-score. In this method, scaling is used to convert floating-point feature values from their natural range into a standard range ($0$ to $1$). The equation is as follows. 
\begin{equation}
\begin{split}
x’ = \frac{x – x_{min}}{x_{max} – x_{min}}
\end{split}
\end{equation}
In this equation, the $x$ is the original value. $x_{max}$ and $x_{min}$ are maximum and minimum values of the feature values. $x’$ is the new value of the feature matrix. When the data is preprocessed and two features are extracted. The result will be used in the SVM model. 

\subsubsection{Support Vector Machine Model Layer}

This layer is based on the support vector machine (SVM) technique to generate a classification model. The SVM model is a supervised machine learning model that uses classification algorithms for two-group classification problems \cite{cortes1995support}. In other words, it analyzes data for regression analysis and classifies the samples into two labels. Given a set of training samples, which are marked as in one of two categories, an SVM method builds a classification model that classifies testing samples into predicted categories and makes a non-probabilistic binary linear classifier. SVM method is also popular in image recognition and classification \cite{okwuashi2020deep,agarap2017architecture}. Its application is broad, just like in agricultural field \cite{islam2017detection}, and the medical field \cite{chowdary2021machine}. In our experiment, we use linear SVM to classify the label. We transform our dataset into the following form.
\begin{equation}
\begin{split}
(\mathbf{x}_1, y_1), …, (\mathbf{x}_n, y_n),
\end{split}
\end{equation}
where the $y_i$ only has two values $0$ or $1$, indicating the class to which the point $\mathbf{x}_i$ belongs. It represents the pruning performance of the fruit tree. Each $\mathbf{x}_i$ is a vector and each dimension is one of the features. SVM model needs to find the maximum margin hyperplane that divides the samples of $\mathbf{x}_i$ into two groups based on the value of $y_i$. The hyperplane is defined as the maximized distance between the hyperplane and the nearest point $\mathbf{x}_i$ from either group.  It can be written as the expression, 
\begin{equation}
\begin{split}
\mathbf{w}^T\mathbf{x} – b = 0,
\end{split}
\end{equation}
where the $\mathbf{w}^T$ is the normal vector to the hyperplane. Then we select two parallel hyperplanes to separate these two classes of data. Additionally, the distance between two hyperplanes is as large as possible. When the distance is maximized, the region bounded by these two hyperplanes is called margin, and the maximum margin hyperplane lies between them. Eventually, the dataset will be taken apart by the maximum margin hyperplane. The one side belongs to the points predicted as good pruning and points on the other side are predicted as not good.

\section{Evaluation}
\label{sec:evaluation}
In this section, the real-world datasets are introduced for evaluation and the process of the experiment that we collect data. Then the application with the dataset into the SVM model and training the model are illustrated in detail. 

A real-world dataset from the entire dataset introduced in Section~\ref{sec:method} is leveraged to validate the OTSU-SVM algorithm. This dataset includes pear trees shadows and over $100$ images of pear trees’ shadows and shapes. The dataset is collected in mid-June 2020, at 11:00 am $-$ 12:00 noon. The location is in Hangzhou, Zhejiang Province, China, whose coordinates are 30.2741° N, 120.1551° E. We use a $50\ cm \times 50\ cm$ whiteboard to collect the shadows. In the experiment, we put the whiteboard down to the pear trees with $50\ cm$ between the board and the branches and leaves. Limited to get the whole shadow of the tree, we randomly select several points under a pear tree to collect the shadow on the whiteboard and take images. Then we extract features of these image data. The mean values of these features are the features of the pear trees. The figure 6 is our process to collect shadows under the pear trees. When the result is obtained, we take images of the collected pear trees to ask experts in the orchard if the pear trees are pruned and the performance of pruning is good. If the performance and production are good, the performance label of the pear tree will be $1$ and set $-1$ when the performance is not good. Eventually, the data of pear trees are preprocessed and used to analyze in the support vector machine model.

\begin{figure}[htbp]
\centerline{\includegraphics[scale = 0.63]{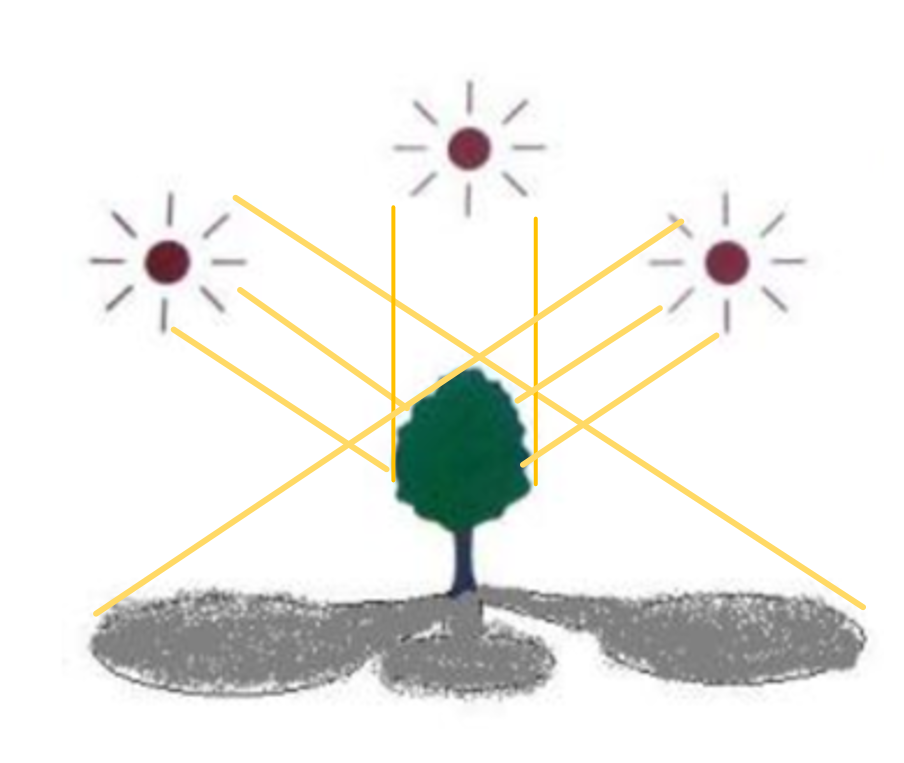}}
\caption{Tree shadows under different degrees of sunlight}
\label{fig9}
\end{figure}

Our experiment design avoids too many issues in the fruit trees. The collection time we select is at noon in June. It is summer in Hangzhou and can make sure the solar zenith angle is close to $90\ degrees$, which can maximize the light area of the fruit tree displayed on the whiteboard and reduce the effect of the shadow of the branch and leaves to the light area. It is shown in Fig.~\ref{fig9}. This figure displays the variation of shadows under the different degrees of sunlight. Smaller degrees of sunlight will lead the shadow to lengthen and features will change out of disproportion. Especially, the shadows of leaves and branches will be stretched and cover the area which is for the transmitted area. Maximizing degrees of sunlight can limit this effect. More importantly, collecting data and images in the same location and same kind of fruit trees can exclude other variables and avoid the effect of other factors which may influence the growth of fruit trees.

\section{Experiment}
\label{sec:experiment}
\begin{figure}[htbp]
\centerline{\includegraphics[scale = 0.4]{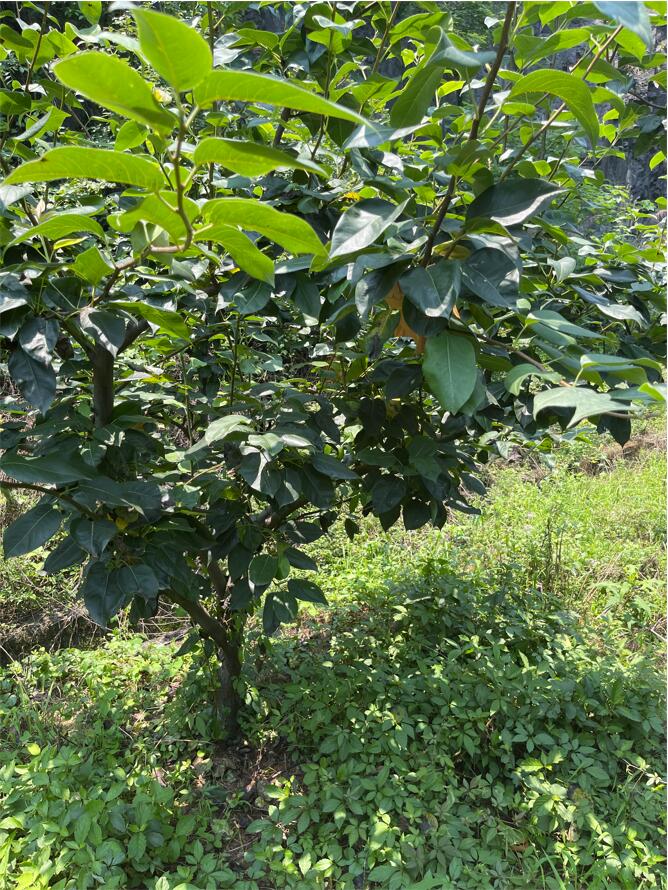}}
\caption{Pear trees in the orchard}
\label{fig10}
\end{figure}
\begin{figure}[htbp]
\centerline{\includegraphics[scale = 0.4]{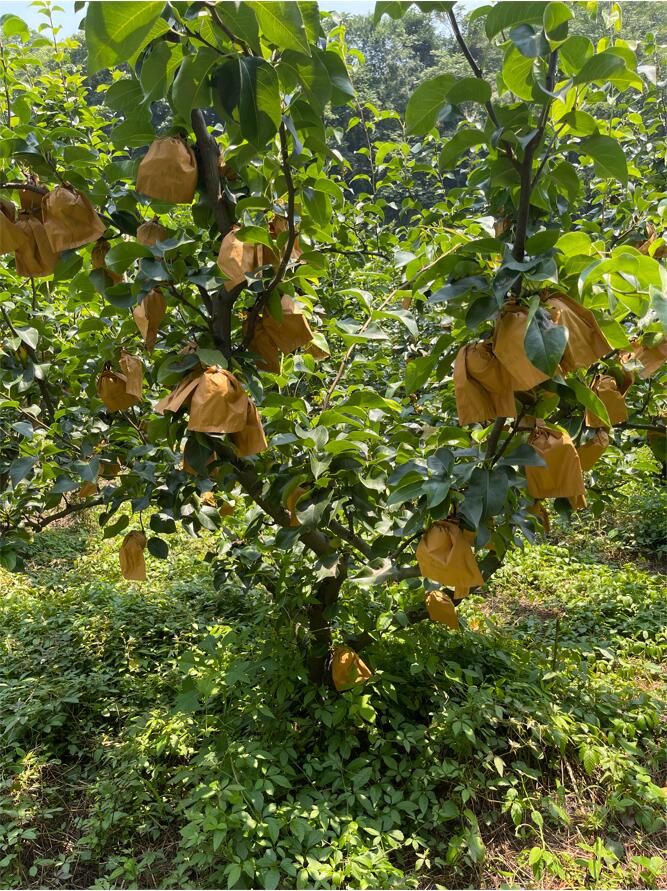}}
\caption{Pear trees in the orchard}
\label{fig11}
\end{figure}
\begin{figure}[htbp]
\centerline{\includegraphics[scale = 0.07]{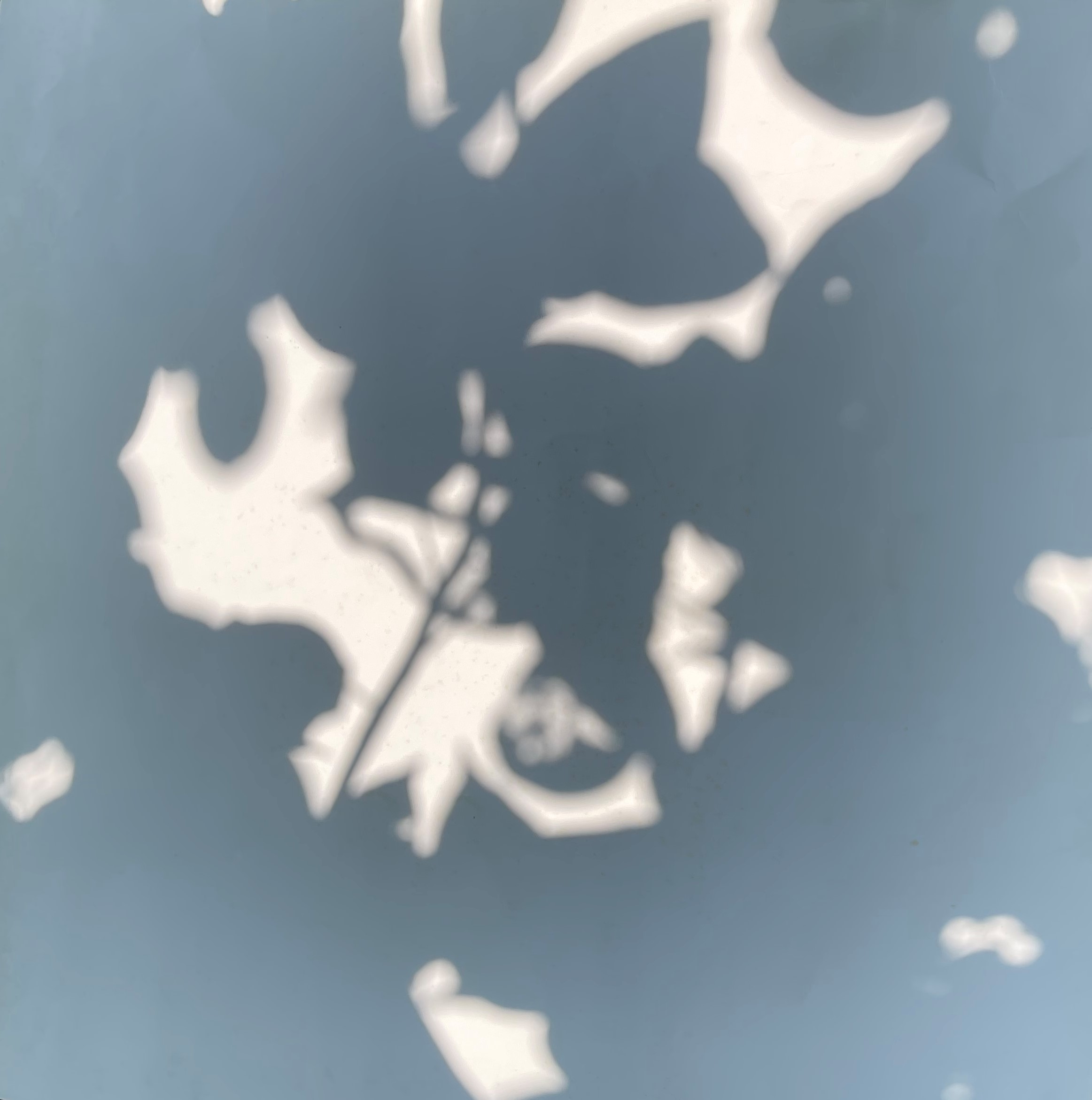}}
\caption{Sample points of pruning not good pear tree}
\label{fig12}
\end{figure}
\begin{figure}[htbp]
\centerline{\includegraphics[scale = 0.07]{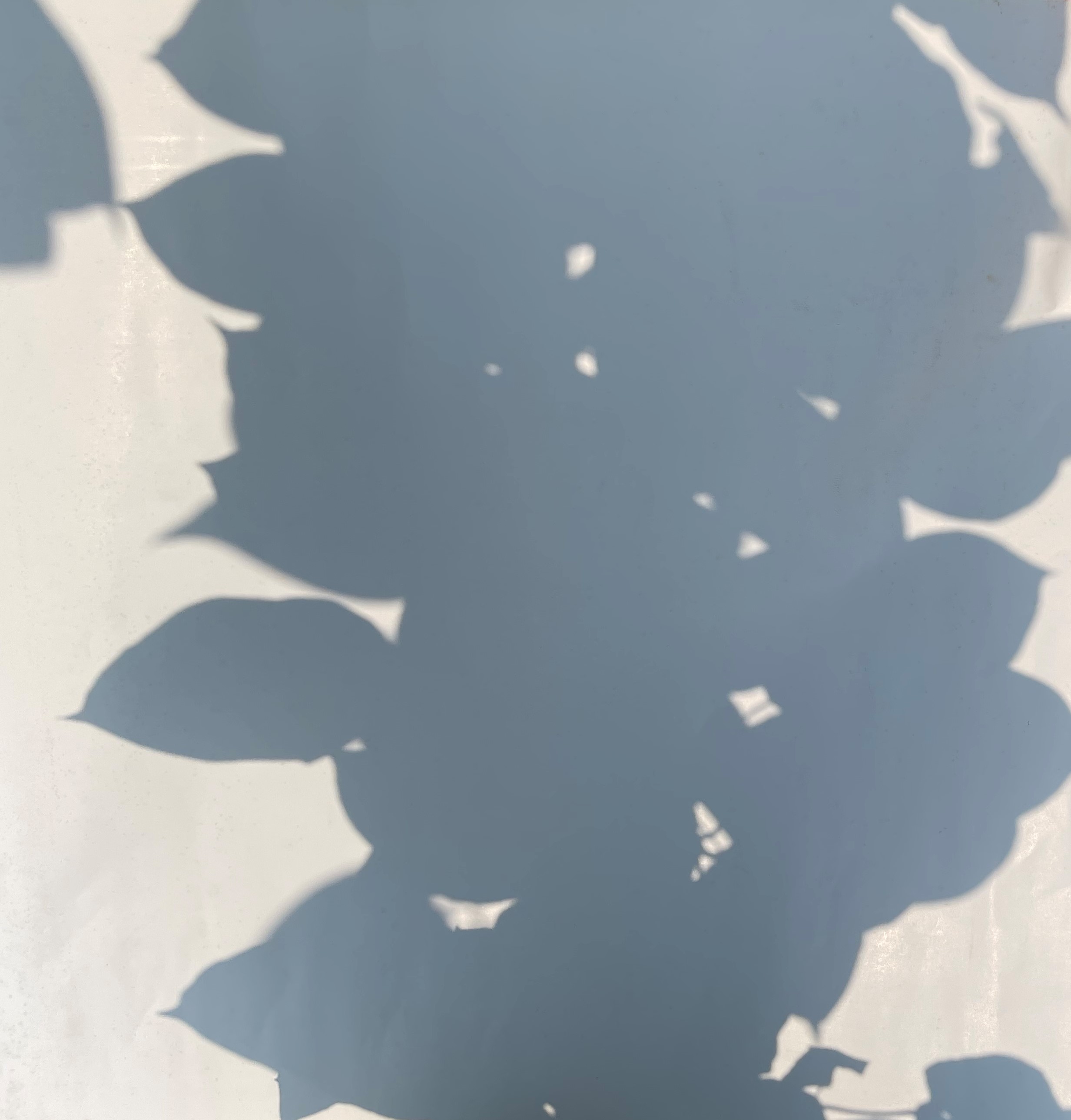}}
\caption{Sample points of pruning good pear tree}
\label{fig13}
\end{figure}

In this section, the process of experiment of this research is introduced in details. The data is collected in mid-June 2020, at 11:00 am ~ 12:00 noon. The orchard is in Hangzhou, Zhejiang Province, China. In this experiment, 100 samples are selected and taken images. These samples are stored into the dataset for pruning trees. Fig.~\ref{fig10} and Fig.~\ref{fig11} are the pear tree selected which are labeled by experts as pruning good or pruning not good.  Fig.~\ref{fig12} and Fig.~\ref{fig13} are the randomly selected points under the trees for shadows. These samples are transformed into matrices and stored into the dataset for pruning trees.
According to the materials and methods section, the method and algorithm of preprocessing and feature selection is followed to implement this classification model and data process in python programing language, which will be upload on the Github website and open source soon. The Numpy library is used to build all matrices of data. It is useful and popular in data processing. Support Vector Machine model is building by sci-kit-learn package, which provides APIs for building the SVM model and generating maximum margin hyperplane to classify the points. Additionally, with purpose to get a clear and obvious result. The Matplotlib package is used to display our result. It is a robust package for computer visualization and data analysis. The input of the model includes the matrix of features called $X$ and the 1-dimension matrix label called $Y$. The table of our data frame is as follows.

\begin{table}[htbp]
\caption{The sample table of data frame including X, Y, and orders}
\begin{center}
\begin{tabular}{|c|c|c|c|}
\hline
\textbf{SUM}&\multicolumn{2}{|c|}{\textbf{X}}&\textbf{Y}  \\
\cline{1-4} 
1 & 0.7531091077884632 & 0.30455985384616746 & 1\\
\hline
2 & 0.6507489189014342 & 0.43863512963846113 & 1 \\
\hline
$\cdots$ & $\cdots$ & $\cdots$ & $\cdots$\\
\hline
9 & 0.6826601194540411 & 0.17369677676678635 & -1\\
\hline
10 & 0.764769075618435 & 0.2990156489318126 & -1\\
\hline
\end{tabular}
\label{tab1}
\end{center}
\end{table}

In this experiment, two methods are used to build the model and predict results. The baseline method is nonlinear SVM with Rbf kernel, which is most common kernel in SVM. The other method is SVM with linear kernel. At first, the dataset are split into two kinds of dataset. The one is training set, the other one is testing set. The proportion of training set and testing set is $0.6$ to $0.4$. The training set is used to build the model and the testing set is used to test the performance of the model. The experiment result is as follows. 
\begin{figure}[htbp]
\centerline{\includegraphics[scale = 0.8]{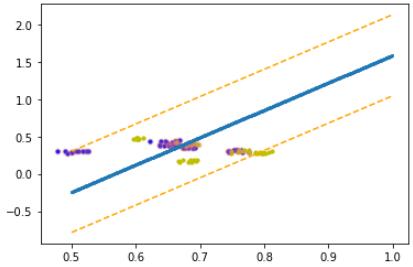}}
\caption{The result of the linear support vector machine model}
\label{fig14}
\end{figure}
\begin{figure}[htbp]
\centerline{\includegraphics[scale = 0.8]{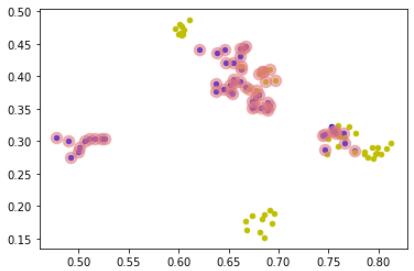}}
\caption{The result of the nonlinear support vector machine model }
\label{fig15}
\end{figure}

Eventually, the output of the model is generated in the following figure. In this figure, the horizon axis is the value of uniformity and the vertical axis means the value of the black pixels rate of the image. These blue points represent the label is 1. In other words, they belong to good pruning fruit trees. The yellow points represent the trees that are pruned not good and label is -1. The light blue line is the hyperplane and it classifies the points into two kinds of pruning performance. The orange dashed lines are the maximum margin of the hyperplane. Additionally, the support vectors are marked and circled. In the They are the points that are the closest to maximum margin hyperplane. 

In the following chart, the accuracy of these two models are shown in these two bars. The nonlinear SVM with Rbf kernel has 47.5$\%$ accuracy and linear kernel has 80.0$\%$. Therefore, in contrast to the baseline method with rbf kernel. Linear kernel SVM has higher accuracy and better performance in the comparison of models.
\begin{figure}[htbp]
\centerline{\includegraphics[scale = 0.3]{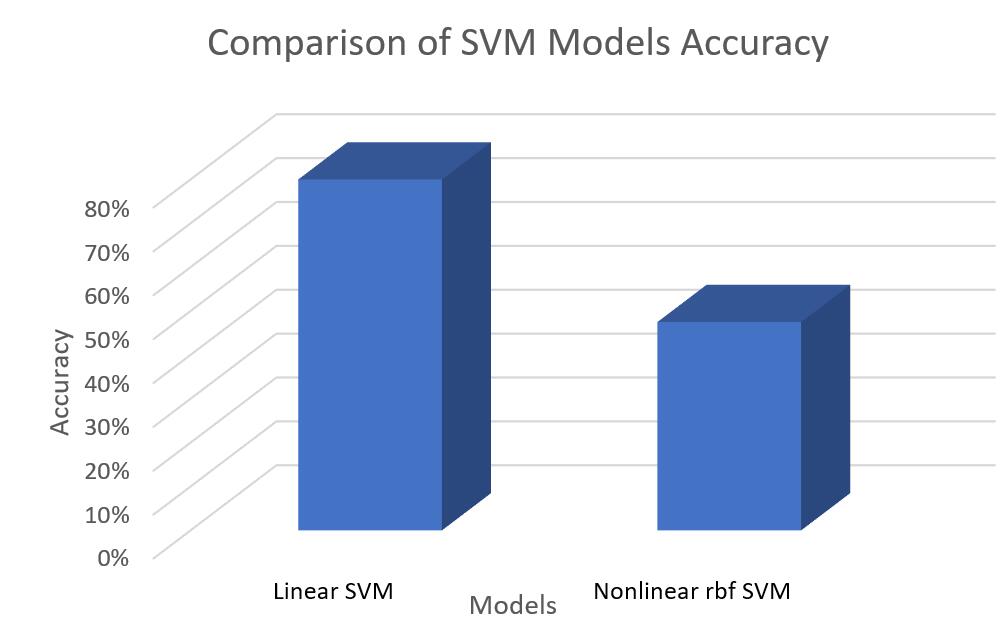}}
\caption{Comparison of SVM Models Accuracy}
\label{fig16}
\end{figure}
\section{Conclusion}
\label{sec:conclusion}
In this paper, the paper designs a support vector machine classification model called OTSU-SVM to evaluate the performance of pruning. It comprehensively measures the features of the fruit trees from two dimensions -- light transmittance dimension and uniformity dimension, and realizes pruning evaluation by integrating these two features and the label of performance.

More importantly, the process of data preprocessing and feature selection plays a significant role before the OTSU-SVM model analysis. We propose three algorithms to preprocess data and measure the feature. The one is to use threshold selection method (OTSU algorithm) to set threshold and determine pixels of the image belong to white or black pixels. The second one is to shrink dimensions and reduce redundancy of the images in order to increase our model efficiency. The last algorithm is to calculate the uniformity of matrix, which is one of features we select. In our method, we define how to measure the black pixels rate and uniformity of shadows. Additionally, we normalize the matrix of features. Normalization is to improve the stability and robustness of the weights of the pruning model. 

The supervised model of machine learning, support vector machine, is used to make a maximum margin hyperplane to classify the good and not good pruning points. Comparing with other pruning decision models, our model has high accuracy and considers more comprehensive factors of the fruit trees. With purpose to prove our model performance, we went to the orchard in Hangzhou, collected the data and took images from pear trees. A dataset is built based on these data and verifies the high accuracy of our model. 

In the future work, we expect these systems can be applied into more areas of agriculture and increase the production of fruits. 

\section{Author Contribution}
Conceptualization, ; methodology, ; software, ; validation,; formal analysis, ; investigation, ; data curation, ; writing—original draft preparation, ; writing—review and editing, ; visualization, ; supervision, ; project administration, ; funding acquisition, . All authors have read and agreed to the published version of the manuscript.
\section{Acknowledgment}
This research is funded by Natural Science Foundation of Zhejiang Province (Grant No. LQ20G010002), and the National Science Foundation of China (71571162, 71702164). This research is supported by the project of China (Hangzhou) cross-border electricity business school, the Philosophy and Social Science Foundation of Zhejiang Province (21NDJC083YB). The authors also gratefully acknowledge Qifeng Yang, which offers us an access to the orchard in Lia Hangzhou, Zhejiang Province, China for collecting data and doing the experiment. He gives us a lot of help on knowledge about the pear trees and pruning.

\bibliographystyle{IEEEtran}
\bibliography{IEEEabrv,reference}

\end{document}